\documentclass[conference]{IEEEtran}
\IEEEoverridecommandlockouts
\usepackage{cite}
\usepackage{amsmath,amssymb,amsfonts}
\usepackage{algorithmic}
\usepackage{graphicx}
\usepackage{textcomp}
\usepackage{xcolor}
\usepackage{bm}
\usepackage{caption} 

\usepackage{comment}
\def\BibTeX{{\rm B\kern-.05em{\sc i\kern-.025em b}\kern-.08em
    T\kern-.1667em\lower.7ex\hbox{E}\kern-.125emX}}
\begin{document}

\title{Learning stochastic dynamical systems with neural networks mimicking the Euler-Maruyama scheme\\
\thanks{This work was supported by the SAD initiative from the Bretagne region and the Carnot TSN institute.}
}
\author{\IEEEauthorblockN{Noura Dridi, Lucas Drumetz, Ronan Fablet}
\IEEEauthorblockA{\textit{(Dept.Mathematical and Electrical Engineering).IMT Atlantique, UMR CNRS Lab-STICC, Brest, France} \\
\{noura.dridi, lucas.drumetz, ronan.fablet\}@imt-atlantique.fr}
}
\maketitle

\def\x{{\mathbf x}}
\def\L{{\cal L}}

\begin{abstract}
Stochastic differential equations (SDEs) are one of the most important representations of dynamical systems. They are notable for the ability to include a deterministic component of the system and a stochastic one to represent random unknown factors. However, this makes learning SDEs much more challenging than ordinary differential equations (ODEs). In this paper, we propose a data driven approach where parameters of the SDE are represented by a neural network with a built-in SDE integration scheme. The loss function is based on a maximum likelihood criterion, under order one Markov Gaussian assumptions. The algorithm is applied to the geometric brownian motion and a stochastic version of the Lorenz-63 model. The latter is particularly hard to handle due to the presence of a stochastic component that depends on the state. The algorithm performance is attested using different simulations results. Besides, comparisons are performed with the reference gradient matching method used for non linear drift estimation, and a neural networks-based method, that does not consider the stochastic term.
\end{abstract}
\begin{IEEEkeywords}
Stochastic differential equations, Maximum likelihood, Neural networks, Lorenz-63 system
\end{IEEEkeywords}

\section{Introduction}
\label{sec:intro}
Understanding the space-time variations of geophysical systems is a challenging task in geosciences. 
Data assimilation approaches typically combine a physical model and remote sensing data or in situ observations. The state-time evolution is represented by a dynamical model, based on an approximate representation of the physics of the real system \cite{Carrassi18}.
The observation model relates the observation to the true hidden state, with a random noise to include observation error and uncertainties.
In this paper, we focus on the state model described by a dynamical model, assuming an identity observation operator.
Dynamical systems are usually represented by Ordinary differential Equations (ODEs) or Stochastic Differential Equations (SDEs).
The latter encompass a noise-driving process in addition to the deterministic component. The noise represents the uncertainty behind the model, i.e. processes not included in the model but present in the real system. They are widely used in domains such as finance\cite{Braumann19}, turbulence study \cite{Friedrich97}, the motion of vehicles in a traffic \cite{Verma16}, oceanography \cite{Brillinger02}.\par
Identification of governing equations of a dynamical system from data can be performed using physical priors and/or machine learning approaches. A method including polynomial representations was proposed in \cite{Paduart10}. In \cite{Brunton16}, the authors propose a sparse regression framework with linear and non linear terms. However, such a formulation cannot deal with SDEs or even observation noise~\cite{Nguyen2020}.
With the availability of larger data sets, machine learning approaches become relevant, using for instance a neural network representation of the unknown operator.
They include autoregressive models such as Recurrent Neural Networks (RNN) \cite{Fablet18}, LSTM \cite{Yeo19}, Resnet \cite{Qin18}, or reservoir computing. The latter has been used successfully for short-term prediction and attractor reconstruction of chaotic dynamical systems from time series data \cite{Pathak17, Lu18}. Besides, the approach provides promising results compared to Echo State Networks (ESNs) for high dimensional chaotic systems. However, good performance is achieved under ideal conditions, i.e. noiseless and regularly sampled with high frequency data.
For dynamical systems represented by ODEs, Neural network based algorithms were proposed \cite{Fablet18},\cite{Ouala19},\cite{Chen18}.
In \cite{Nguyen2020}, \cite{Bocquet20} the identification used noisy and partial observations, therefore both estimation of the hidden state and identification of governing equations are tackled.
For SDEs, the identification comprises estimation of the drift and the diffusion. It can be performed using pre-defined parametric representations of the drift and the diffusion \cite{Friedrich11}.
Non-parametric representation of the drift includes linear and non linear drift. The first can be carried using a variational smoothing algorithm \cite{Archambeau07} or a variational mean
field approximation \cite{Vrettas15}. Non linear drifts are modelled using Gaussian processes \cite{Yildiz18}. For both parametric and non parametric drifts, the gradient matching approximation method is largely exploited \cite{Dony19}, \cite{Niu16}.
The drift is estimated to match the
empirical gradients of data, while the
diffusion relates to the residual of the approximation.
On the other hand, in \cite{Li20}, the authors generalize the adjoint sensitivity approach to compute gradients of the solutions and combine with a stochastic variational inference scheme to train a latent SDE. Besides, the method is used to the Lorenz-63 system~\cite{Lorenz63}, with an additive diffusion term linearly dependent on the state. Multiple trajectories are required for learning.\par
In this work we propose to extend \cite{Fablet18} to model stochastic dynamical systems. Although our motivation and applications are related to problems in ocean and atmospheric sciences, the method can be applied for more general issues.
 The method is used to learn the Geometric Brownian Motion (GBM) parameters. The GBM is a reduced dimension toy model, that includes a stochastic component depending on the state.
A second application is to learn the dynamical operator of a stochastic version of the Lorenz 63 model.
The deterministic Lorenz model introduced in \cite{Lorenz63} is a system of three coupled ODEs including linear and non linear terms. The Lorenz-63 system is an appropriate simplified representation of the ocean-atmosphere interactions derived from the Navier-Stokes equations.
The Stochastic Lorenz (SL) system presented in \cite{Chapron18}
describes the evolution of the deterministic Lorenz model when the local position of the state is uncertain. 
This uncertainty is represented by a multiplicative noise on the $y$ and $z$ components representing variables induced by small scale velocity fluctuations. $x$ is the large scale variable, which is indirectly affected by the small scale uncertainty.
The SL system is an important model to represent geophysical flows. It helps to efficiently explore the entire dynamical landscape of the flows, with a reduced order flow representation \cite{Chapron18}, \cite{Yang19}.
We propose a deep learning based approach to identify the parameters of an SDE, in which both the drift and diffusion are represented using neural networks. A parametric formulation is considered, without prior assumptions on the parameters.
They are estimated using a maximum likelihood criterion, to define the loss function of the learning algorithm. We maximize the probability of obtaining the observed time series with a Markovian model of order one on the time samples.
The performance are attested on the GBM and the SL system.
Compared with an algorithm based on a deterministic formulation, the simulations results illustrate, the relevance of considering an SDE formulation to deal with the SL system.
The remainder of the paper is organized as follows: section \ref{sec:NN} is dedicated to the neural networks for stochastic dynamical system, section \ref{sec:LA} describes the learning algorithm. Simulations results are presented in section \ref{sec:Res}. We end the paper by a conclusion with the perspectives of this work.
\section{Neural Networks for stochastic dynamical systems}
\label{sec:NN}
In this section the Neural Network (NN) to represent and generate a process using a stochastic differential equation, is described. A SDE takes the general form of
\begin{equation}\label{eq:SDE}
d\bm x=\mathcal{F}(\bm x(t),t)dt+\mathcal{L}(\bm x(t),t) d\bm \beta_t
\end{equation}
where $\mathcal{F}$ and $\mathcal{L}$ are linear/non linear functions of $\mathbf{x}$ and $d\bm \beta_t$ is a multivariate Brownian motion, $\bm x$ is a $d$-dimensional vector, $\mathcal{F}: \mathbf{R}^d\times \mathbf{R} \rightarrow \mathbf{R}^d $ is a function of $\bm x$ and the time $t$, $\mathcal{L}$ is a state dependent matrix in $\mathbf{R}^{d\times s}$, and $d\bm \beta_t$ a vector of dimension $s$.
For one dimensional case, $d=s=1$ and linear functions $\mathcal{F}$ and $\mathcal{L}$, an analytical solution of the SDE can be calculated such as with the Ornstein–Uhlenbeck process or the GBM. However, in general, an explicit solution cannot be obtained from Eq(\ref{eq:SDE}) and numerical approximation methods are used, such as the Euler-Maruyama (EM) integration scheme:
\begin{equation}\label{eq:EM}
\bm x_{t+\tau}= \bm x_{t}+\tau\mathcal{F}(\bm x_t,t)+\mathcal{L}(\bm x_t,t) \bigtriangleup\beta_t
\end{equation}
with $\tau$ is the time resolution. 
The EM method is the equivalent to Euler integration scheme for SDEs \cite{Kloeden92}.
The EM can be used to simulate trajectories from SDEs and the result converges to the true solution in the limit $\tau \rightarrow 0$. When $\mathcal{L}$ equals to identity, the formulation given by Eq.(\ref{eq:EM}), can be regarded as an autoregressive model of first order.\\
Hereafter, unless otherwise specified, $\tau$ is set to 1 to simplify the notation, without any loss of generality.
Considering a neural network architecture, the EM approximation Eq.(\ref{eq:EM}) can be regarded as a recurrent network with one residual layer, and the SDE in Eq.(\ref{eq:SDE}), is reformulated as:
\begin{equation}\label{eq:NN}
d\bm x= \left(A_1\bm x(t)+A_2\bm x(t)\times A_3\bm x(t)\right)dt+B\bm x(t)d\beta_t
\end{equation}
where $A_i$, for i in ${1,2,3}$ and $\textit{B}$ are scalar valued matrices representing respectively the operators $\mathcal{F}$ and $\mathcal{L}$.
The architecture of the network is defined based on Eq(\ref{eq:NN}), so that the formulation of the neural network includes the true parameterization of the model.
This is a general architecture that is convenient for the GBM an the SL models.
The first term corresponds to the linear component of the model represented by a fully connected layer, the second term includes bilinear elements of the model represented by an element-wise product operator to embed a second-order polynomial representation for operator $\mathcal{F}$. 
The whole representation for the first two components is a bilinear neural net architecture as in  \cite{Fablet18}. The third term refer to the stochastic part of the model with multiplicative noise, where the linear component is designed for $\mathcal{L}$.
The network is represented with a fully connected layer multiplied by sampling layer that simulates the noise.
This sampling layer is used only for the model to be able to generate new trajectories at test phase, but is not directly used in the training phase since the training is not stochastic.
\vspace*{-0.9cm}\\
\section{Learning algorithm}
\label{sec:LA}
\vspace*{-0.3cm}
The aim is to identify the unknown SDE, that is to estimate $\mathcal{F}$ and $\mathcal{L}$ from the data.
Based on the NN representation, the identification can be stated as learning the parameters of a recurrent residual network.
The deterministic dynamical system is represented as an ODE, and a learning algorithm using a residual neural network architecture was proposed in \cite{Fablet18}. The latter approach is related to the Neural ODE scheme that has recently gained popularity \cite{Chen18}, the authors propose an ODE solver using continuous-depth models.
A theoretical guarantee of convergence was proved in \cite{Chen18} and \cite{Li20}, showing that residual networks are well suited to learn from data governed by both ODEs or SDEs.\par
In this paper, we propose to extend the architecture to learn the stochastic dynamical model from data. Given an initial condition, the EM method Eq(\ref{eq:EM}) is used to generate the training trajectory.
The goal is to identify the set of parameters $\Theta=\{\omega, \phi \}$ from the data, where $\omega$ and $\phi$ denote the parameters of $\mathcal{F}$ and $\mathcal{L}$, respectively. To relate with the NN notation in Eq.(\ref{eq:NN}), $\omega=\{A_1,A_2,A_3\}$, and $\phi=\{B\}$. It is worth pointing here that the stochastic nature of the model complicates the reconstruction process. Indeed, even for known parameters we cannot reconstruct the original trajectory used for training. Therefore, unlike for deterministic systems \cite{Fablet18}, the short-term prediction error is not suitable as optimization criterion.
From a probabilistic point of view, and given that the state process is Markovian, the likelihood function is given by:
\begin{equation}\label{eq:lik}
    p_{\Theta}(\bm x_{0:T})=p_{\Theta}(\bm x_0)\prod_{t=0}^T p_{\Theta}(\bm x_{t+1}|\bm x_t)
\end{equation}
where $\textit{T}$ is the number of points in the trajectory. From Eq.(\ref{eq:EM}), the distribution of $\bm x_{t+1}$ given $\bm x_{t}$  is obtained:
\begin{equation}\label{eq:disx}
    \bm x_{t+1}|\bm x_t \sim N(\bm m_t,\bm \Sigma_t)
\end{equation}
where $N$ refers to a multivariate Gaussian distribution with mean $\bm m_t=\bm x_t+\mathcal{F}(\bm x_t)$,
and covariance matrix $\bm \Sigma_t= \mathcal{L}(\bm x,t)\mathcal{L}^T(\bm x,t)$. It is worth pointing here that only the the conditional distribution $\bm x_{t+1}|\bm x_t$ is Gaussian and not the whole process. 
Plugging Eq.(\ref{eq:disx}) into Eq.(\ref{eq:lik}), and considering $\mathcal{F_\omega}$ and $\mathcal{L_\phi}$, then the maximum likelihood estimation of the parameters is equivalent to the minimization:
\begin{equation}\label{eq:loss}
  \arg \min_{\omega,\phi} \sum_{t=0}^{T}\|\bm x_{t+1}-\bm m_t\|_{\bm \Sigma_t^{-1}}^2+\sum_{t=0}^{T} \log|\bm \Sigma_t|
\end{equation}
Eq.(\ref{eq:loss}), defines the loss function used to train the NN. Unlike the deterministic case, $\bm \Sigma_t$ can not be reduced to the identity matrix to obtain the short term prediction error as in \cite{Nguyen2020}.
\section{Results}
\label{sec:Res}
Numerical experiments are presented to evaluate the performance of the proposed algorithm. Data are generated using the EM scheme Eq.(\ref{eq:EM}), for fixed parameters and time steps. The proposed algorithm named Bi-NN-SDE is compared to:
\begin{itemize}
    \item The deterministic algorithm proposed in \cite{Fablet18}, using residual Bilinear Neural Network architecture, named Bi-NN.
    \item A Gradient matching approximation predominantly used for non linear drift and diffusion estimation \cite{Friedrich11}.
    Drift and the diffusion are respectively estimated using an ensemble average for non stationary processes.
    \begin{eqnarray}\label{eq:GM}
    \hat{\mathcal{F}}(\bm x_t)&=& \frac{1}{ N} \sum_{t_j \in \alpha} \{\bm x(t_j+1)-\bm x(t_j) \} \\
     \hat{\mathcal{L}}(\bm x_t)&=& \frac{1}{N} \sum_{t_j \in \alpha} \{\bm x(t_j+1)-\bm x(t_j)- f(\bm x_{t_j},\bm x_j) \}^2\nonumber 
    \end{eqnarray}
    \textit{N} is the number of the ensemble realizations used for the stochastic process, one realization is denoted by $\alpha$.
\end{itemize}
First, the algorithm is applied to learn the parameters of an SDE that describes a GBM process $x(t)$ \cite{Ross14}:
\begin{equation}
    dx=\mu xdt+\sigma x d\beta_t
\end{equation}
Our goal is to estimate the drift $\mathcal{F}=\mu$ and the volatility $\mathcal{L}=\sigma$ then reconstruct the trajectories using the learned model from a given initial condition.
For the GBM, the theoretical mean and variance are known:
\begin{eqnarray}\label{eq:mom_gbm}
E[x]&=&x_0\exp(\mu t)\nonumber\\
V(x)&=&x_0^2\exp(2\mu t)[\exp(\sigma^2t)-1]
\end{eqnarray}
These values are compared with their equivalent by replacing $\mu $ and $\sigma$ by the learned values.
Comparison includes also, empirical curves measured using 1000 trajectories generated respectively by the true model and the learned ones using the algorithm BiNN and BiNN-SDE. Below, we consider $\mu=0.5$ and $\sigma=1$, $\tau=0.001$ and the trajectory length $T=3000$.
\vspace{-0.2cm}
 \begin{figure}[h]
\centering
\includegraphics[width=8cm,height=3cm]{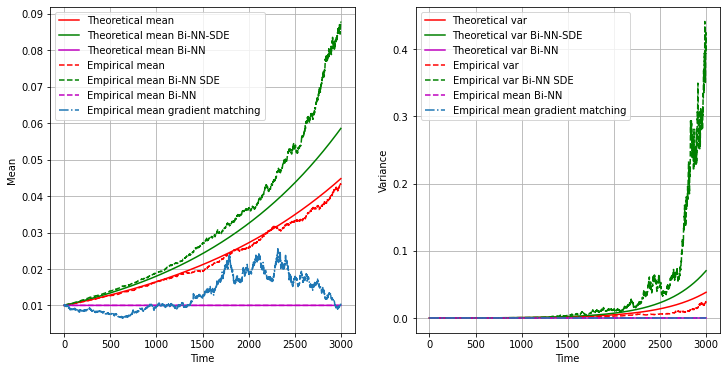}
\caption{\textit{Theoretical and empirical mean and variance for the GBM process, with $\mu=0.5$, $\sigma=1$}}
\label{fig:mean_cov_gama1mu05_2}
\end{figure} 
Fig \ref{fig:mean_cov_gama1mu05_2} represents one representative run of the model, corresponding to a learned drift $\hat{\mu}$ and a diffusion $\hat{\sigma}$ close to the mean value obtained over $100$ trainings using different trajectories table.(\ref{tab:param_est}). 
Curves with solid lines correspond to the theoretical mean and variance, while the one with dashed lines refer to the empirical parameters. The red color corresponds to the true model, while the green, purple and blue are assigned respectively to the BiNN-SDE, BiNN and the gradient matching. For the latter, since we don't have access to the parameters, there are no curves for the theoretical mean and variance.
It is illustrated that, despite the stochastic nature of the model, the Bi-NN-SDE algorithm learns the mean and variance of the GBM, with lower error than the Bi-NN and the gradient matching. For $t<2000$, performance of the proposed algorithm are slightly different from the theoretical one, and it is expected to increase for higher $t$.
For the BiNN algorithm, the learned trajectory is deterministic, therefore theoretical and empirical parameters are almost the same.\par
On the other hand to evaluate the variability of the estimator, we present the mean and variance of the learned parameters using 100 different learned models table.(\ref{tab:param_est}), confirming the superiority of the BiNN-SDE algorithm. Note that with the Bi-NN a deterministic formulation is considered, therefore the drift $\sigma$ is not estimated. The gradient matching algorithm, provides an approximation of $\mathcal{F}(x_t)$ and $\mathcal{L}( x_t)$, thus an explicit form of $\mu$ and $\sigma$ cannot be calculated.\\
\begin{table}[!h]
\centering
\begin{tabular}{|p{1.4cm}|p{1.7cm}|p{1.7cm}||p{1.7cm}|} 
  \hline
Algorithm & Bi-NN-${\mu}$& Bi-NN-SDE-${\mu}$&Bi-NN-SDE-${\sigma}$\\
   \hline
Mean  & $0.188$& $0.535$&$0.998$\\
 \hline
Variance & $ 0.083$& $0.201$&$ 10^{-4}$\\
 \hline
\end{tabular}
\caption{\textit{Mean and variance of the learned drift (second and third columns) and the learned diffusion (fourth column).The true values equal to $\mu=0.5$ and $\sigma=1$}}
\label{tab:param_est}
\end{table}
Second, the algorithm is applied to the SL system, that is a stochastic version of Lorenz-63 model. 
The latter is a dynamical system largely used to
represent ocean-atmosphere interactions: it presents chaotic patterns, i.e. a small perturbation in the initial condition leads to very different trajectories in the long run. 
Moreover, the Lorenz-63 is a nice toy model to learn dynamics from data : it is a reduced order, easy to visualise model including non linear interactions.
The stochastic version of the Lorenz system, introduced in \cite{Chapron18} is applied to represent large-scale geophysical flows, with uncertainty behind the local position of the state. 
We propose to model the system as an SDE and thus the identification includes the estimation of the drift and the diffusion terms. Identification of such a system is  difficult due to the multivariate dimension, non linear relations between state components, the chaoticity of the system in addition to the stochastic component. Besides, it is worth pointing that the diffusion term depends on the state, that is different from a simple deterministic Lorenz with additive noise.
The model is given as follows:
\begin{eqnarray}\label{eq:LS}
\frac{dx}{dt}&=& \sigma y-\left(\sigma+\frac{2}{\gamma}\right)x\label{eq:LS_1}\\
dy&=&\left( \left(\rho-z \right)x-\left(1+\frac{2}{\gamma} \right)y\right)dt+\frac{\rho-z}{\sqrt{\gamma}}d\beta_t\label{eq:LS_2}\\
dz&=&\left(xy-\left(\beta +\frac{4}{\gamma}\right)z\right)dt+\frac{y}{\sqrt{\gamma}} d\beta_t\label{eq:LS_3}
\end{eqnarray}
$d\beta_t$ is a Brownian motion, and $\gamma$ characterizes the noise level: the higher $\gamma$, the lower the noise level. $\sigma$, $\rho$ and $\beta$ are the system parameters, also present in the deterministic version of the Lorenz 63.
The model is composed of an ODE on the velocity Eq.(\ref{eq:LS_1}), with two SDEs associated to temperature fluctuations Eq.(\ref{eq:LS_2}) and Eq.(\ref{eq:LS_3}).\\
Note that for low noise level $\gamma \rightarrow \infty$, we retrieve the deterministic Lorenz equations\cite{Lorenz63}, and that the $\gamma$ influences the $x$ variable, though it is not directly perturbed by noise.
The SL model can be considered as stochastic differential equations that take the general form of Eq.(\ref{eq:SDE}).
Note the difference with the GBM, the dimension (for the SL it equals to 3), that implies higher number of parameters to learn: 27 for $\mathcal{F}$ and 9 for $\mathcal{L}$.
The evaluation is performed on estimation of the parameters using the RMSE, by matching the obtained parametrization in Eq.(\ref{eq:NN}) with the true one (including the coefficients that should be zero). The ability of the algorithm to reconstruct the attractor of the SL system is also illustrated. 
The data are generated using the EM method from the true SL model, for one trajectory, a fixed number of time steps and an integration step, and different noise levels.\\
We simulate $N_s=100$ trajectories of the SL model, with $\tau=10^{-3}$, and a trajectory length of $T=10000$, $\gamma=50$ (low noise level) and $\gamma=10$ (high noise level). The models are trained once on each individual trajectory.
\begin{table}[!htb]
\centering
\begin{tabular}{|p{1cm}|p{0.8cm}|p{0.8cm}||p{0.8cm}|p{0.8cm}|p{0.8cm}|p{0.8cm}|} 
  \hline
Algorithm &  \multicolumn{2}{c||}{Bi-NN-${\mathcal{F}}$}&\multicolumn{2}{c|}{Bi-NN-SDE-${\mathcal{F}}$}& \multicolumn{2}{c|}{Bi-NN-${\mathcal{L}}$}\\
   \hline
   Noise level $\gamma$ & $50$&$10$& $50$&$10$& $50$&$10$\\
   \hline
Mean RMSE  & $0.322$& $0.601$& $0.306$&$0.522$&$0.123$ & $0.359$\\
 \hline
Var RMSE& $0.019$& $0.065$&$0.017$& $0.0447$& $0.002$& $0.012$\\
 \hline
\end{tabular}
\caption{\textit{Mean and variance of the RMSE}}
\label{tab:rmse_gama50_10}
\end{table}

Table (\ref{tab:rmse_gama50_10}) shows the mean and variance of the RMSE for the estimation the drift and the diffusion matrices. For $\gamma=50$, the RMSE provided by the Bi-NN-SDE is lower than the Bi-NN, but all models perform relatively well, due to the small noise level. Furthermore, for strong noise $\gamma=10$, using the Bi-NN-SDE algorithm clearly improves the estimation performance. The difference between RMSE is multiplied by five ($0.016 \rightarrow 0.08$). The above results present the global RMSE. However, the parameters of the model encompass a large set of zero coefficients. Therefore, it is convenient to give the estimation error for the non-null coefficients, and to check that the zero coefficients are close to zero. 
 \begin{figure}[!htb]
\centering
\includegraphics[width=8cm,height=3.5cm]{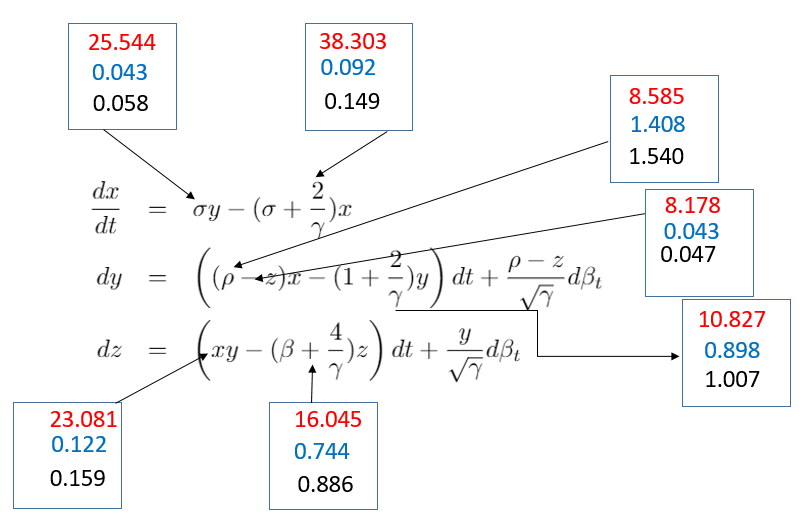}
\caption{\textit{Mean of the RMSE per coefficient, black numbers refer to the Bi-NN, blue to Bi-NN-SDE and red for the gain rate.}}
\label{fig:rmse_modele}
\end{figure}
The gain rate is computed by the difference between the RMSE of both algorithms normalized by the RMSE of the Bi-NN.
As shown in Fig(\ref{fig:rmse_modele}), the Bi-NN-SDE outperforms the Bi-NN one for all the model coefficients, the gain rate obtained is higher than $8\%$ and can attain $38\%$.
Regarding the null coefficients, the individual RMSE  is $\leq 10^{-2}$ except for 3 coefficients (from 20 null ones) where it is of order of $10^{-1}$.\par
Another possible parameterization involves setting to zeros the null coefficients of the diffusion term to reduce the number of parameters to learn and therefore improve the performance. This is physically acceptable, since the first equation of SL representing the large scale variable is deterministic. Due to limited space, results are not presented for this formulation.\par
\begin{figure}[!htb]
\centering
\includegraphics[width=7cm,height=3cm]{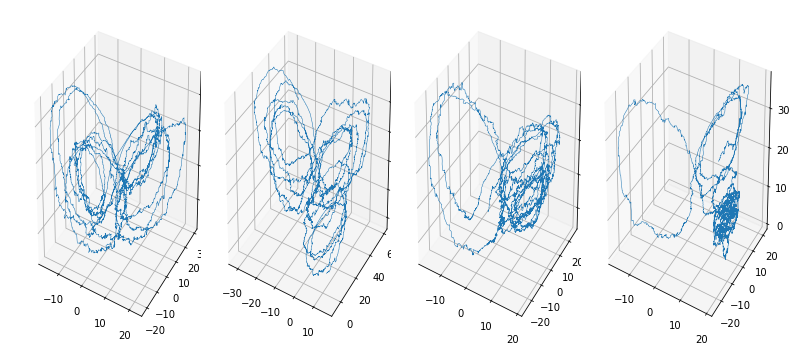}
\caption{\textit{SL trajectories generated using the gradient matching.}}
\label{fig:exp_gradmatch}
\end{figure}

\textbf{Comparison of the obtained attractors}\\
We propose to further illustrate the algorithm performance with visual comparison of the attractors generated by the true model and the one using the different learning algorithms. A more quantitative comparison is challenging due to the stochasticity: for instance extending Lyapunov exponents (which quantify the chaotic nature of attractors) to stochastic systems is not straightforward.
 \begin{figure}[!htb]
\centering
\includegraphics[width=7cm,height=5cm]{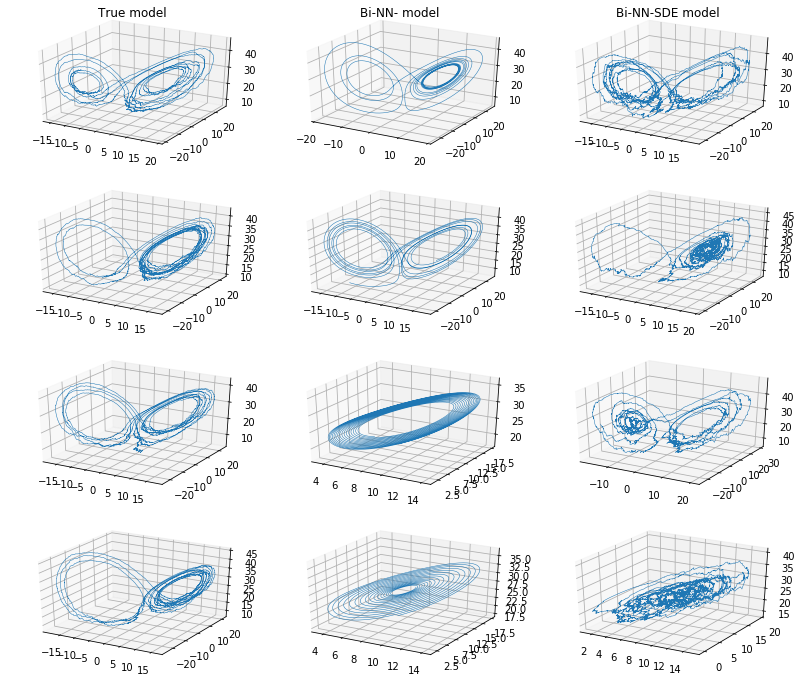}
\caption{\textit{Different results for the SL trajectories generated using the true model and the learned ones with Bi-NN and Bi-NN-SDE, respectively}}
\label{fig:attract}
\end{figure} 
We observe in Fig(\ref{fig:attract}) that among the $N_s=100$ experiments with $\gamma=10$, in 
$90\%$ of cases (results of the first and third row) the sequence generated by the learned model Bi-NN-SDE captures the topology of the true model (the two side of the butterfly shape of the attractor), while the Bi-NN  only achieves that in $60\%$ of the cases (results showed in the first row), and is stuck only in one side in other cases. Another major difference is that the Bi-NN algorithm fails to detect the stochastic fluctuations since the algorithm consider the representation of the system as an ODE.
Regarding the gradient matching method, given the estimators $\hat{\mathcal{F}}$ and $\hat{\mathcal{L}}$ Eq(\ref{eq:GM}), different attractors are generated for different trajectories used to perform estimation. Some examples are given in Fig(\ref{fig:exp_gradmatch}),
as observed, while the learned system is indeed stochastic, the gradient matching method fails to preserve the topology of the SL model.
In summary Bi-NN-SDE, achieves both recovery of the SL attractor and the stochastic fluctuations in most of the cases, even for significant noise levels.
\section{Conclusion}
\label{sec:Conc}
An algorithm to learn stochastic dynamical systems is proposed.
 The system is modelled as an SDE, the drift and diffusion matrices are parameterized using NN. The algorithm mimicks the EM integration scheme, thus coming with an embedded SDE integration scheme. The networks are trained using a maximum likelihood on the posterior one-step ahead density of the samples. Preliminary results on the GBM process and a challenging SL model illustrate the relevance of considering an SDE formulation and good estimation performance in term of the RMSE.
 We are currently working on improving the quantitative performance of the algorithm, especially for the diffusion term, and extending the algorithm to include stochastic training as in variational autoencoders.
\bibliographystyle{IEEEbib}
\bibliography{main_v6}

\end{document}